\documentclass[sigconf]{acmart}

\AtBeginDocument{%
 }

\usepackage{newtxtext}

\usepackage{graphicx}
\usepackage{subfigure}

\usepackage{algorithm}
\usepackage[noend]{algpseudocode}

\usepackage{tikz}
\usetikzlibrary{fadings}
\usetikzlibrary{shapes,snakes}

\usepackage{url}
\usepackage{multirow}
\usepackage{verbatim}

\usepackage{balance}
\usepackage{xcolor}

\usepackage{hyperref}
\usepackage{stfloats}

\def\lc{\left\lceil}   
\def\rc{\right\rceil}

\def\naive{na\"{\i}ve }

\copyrightyear{2024}
\acmYear{2024}
\setcopyright{rightsretained}
\acmConference[KDD '24]{Proceedings of the 30th ACM SIGKDD Conference on Knowledge Discovery and Data Mining}{August 25--29, 2024}{Barcelona, Spain}
\acmBooktitle{Proceedings of the 30th ACM SIGKDD Conference on Knowledge Discovery and Data Mining (KDD '24), August 25--29, 2024, Barcelona, Spain}\acmDOI{10.1145/3637528.3671521}
\acmISBN{979-8-4007-0490-1/24/08}

\makeatletter
\gdef\@copyrightpermission{
  \begin{minipage}{0.3\columnwidth}
   \href{https://creativecommons.org/licenses/by/4.0/}{\includegraphics[width=0.90\textwidth]{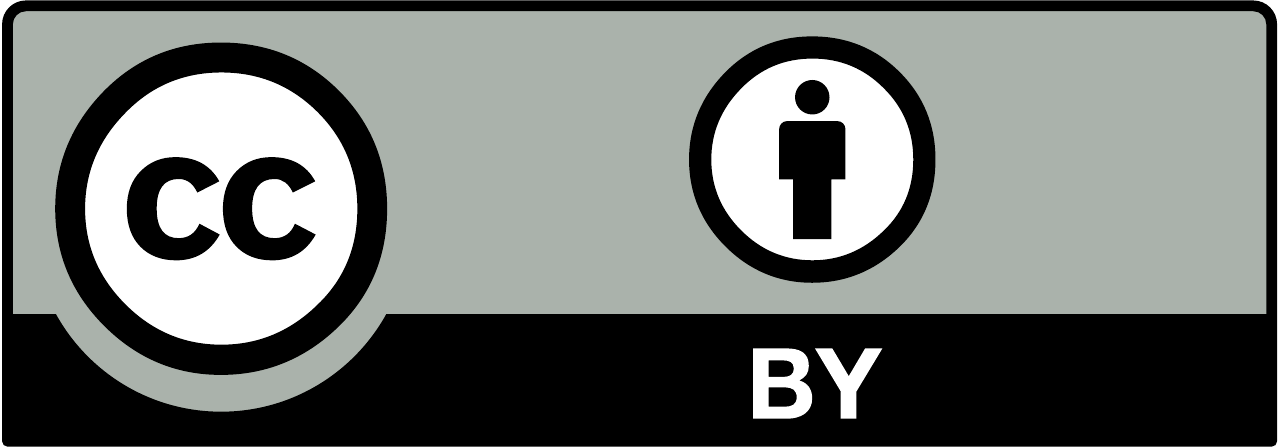}}
  \end{minipage}\hfill
  \begin{minipage}{0.7\columnwidth}
   \href{https://creativecommons.org/licenses/by/4.0/}{This work is licensed under a Creative Commons Attribution International 4.0 License.}
  \end{minipage}
  \vspace{5pt}
}
\makeatother

\begin{document}

\title{On Finding Bi-objective Pareto-optimal Fraud Prevention Rule Sets for Fintech Applications}

\author{Chengyao Wen}
\orcid{0009-0009-3928-4235}
\affiliation{
  \institution{Ant Group}
  \city{Chengdu}
  \country{China}
}
\email{wenchengyao.wcy@antgroup.com}

\author{Yin Lou}
\orcid{0009-0004-8329-7192}
\affiliation{
  \institution{Ant Group}
  \city{Sunnyvale, CA}
  \country{USA}
}
\email{yin.lou@antgroup.com}

\begin{abstract}
Rules are widely used in Fintech institutions to make fraud prevention decisions, since rules are highly interpretable thanks to their intuitive if-then structure. In practice, a two-stage framework of fraud prevention decision rule set mining is usually employed in large Fintech institutions; Stage 1 generates a potentially large pool of rules and Stage 2 aims to produce a refined rule subset according to some criteria (typically based on precision and recall). This paper focuses on improving the flexibility and efficacy of this two-stage framework, and is concerned with finding high-quality rule subsets in a bi-objective space (such as precision and recall). To this end, we first introduce a novel algorithm called SpectralRules that directly generates a compact pool of rules in Stage 1 with high diversity. We empirically find such diversity improves the quality of the final rule subset. In addition, we introduce an intermediate stage between Stage 1 and 2 that adopts the concept of Pareto optimality and aims to find a set of non-dominated rule subsets, which constitutes a Pareto front. This intermediate stage greatly simplifies the selection criteria and increases the flexibility of Stage 2. For this intermediate stage, we propose a heuristic-based framework called PORS and we identify that the core of PORS is the problem of solution selection on the front (SSF). We provide a systematic categorization of the SSF problem and a thorough empirical evaluation of various SSF methods on both public and proprietary datasets. On two real application scenarios within Alipay, we demonstrate the advantages of our proposed methodology over existing work.
\end{abstract}



\maketitle

\section{Introduction}
\label{sec:intro}

Fraud prevention is an important task in Fintech applications. Fraudulent activities usually include identity theft, money laundering, fraudulent payment transactions, etc. Whenever a fraud is committed, the loss is not only incurred by the victim who is exploited, but the reputation of the financial institution involved also takes a hit. 

In many Fintech fraud prevention applications, interpretability is usually a must-have requirement in addition to predictive accuracy. Therefore, rule-based models are widely used for such applications to make fraud prevention decisions~\cite{li2022adaptive}, since rules can offer intuitive representation of knowledge, thanks to their simple if-then structure.\footnote{It is important to note that we do not exclude ourselves from using more advanced machine learning models; signals from those models (along with all other more interpretable features, such as user profile information) are input features to rule learning algorithms. Due to the regulatory requirement, even if the decision is from a DNN model, it should always have the form ``if model\_score > threshold then 1.''}  In practice, decision rule sets (also known as disjunctive normal form) are often favored over decision rule list due to their flatter representation~\cite{lakkaraju2016interpretable, li2022adaptive}; rule set models make positive prediction for a transaction whenever any rule in the set is satisfied. During any stage in a transaction, whenever an alert is made from the rule set (at least one rule predicts positive on this transaction), the transaction is interrupted and the user may be asked to submit additional information to verify the legitimacy for this transaction.

In practice, a two-stage framework of rule set mining is commonly employed in Fintech institutions (e.g., Alipay\footnote{\url{https://www.alipay.com/}})~\cite{li2022adaptive}. In Stage 1, a potentially large pool of rules is generated. Stage 2 aims to produce a refined rule subset according to some criteria (typically based on precision and recall). Note that in this paper we use rule set to refer to rules generated by Stage 1 and rule subset to refer to a refined subset of the initial pool of rules \emph{after} Stage 1.  

The state-of-the-art approach to Fintech fraud prevention applications employs tree ensemble in Stage 1~\cite{li2022adaptive}. We empirically find such tree-based approach can lead to unnecessarily large pool of ``homogeneous'' rules (in terms of precision and recall) in Stage 1 (See Section~\ref{sec:sr_vs_tree_ens}), which subsequently hurts both the efficiency and the quality of the refined rule subset for Stage 2. As noted in~~\cite{li2022adaptive}, various filters are needed to reduce the number of rules produced in this stage for better efficiency and effectiveness. Therefore, in this paper we first propose a novel algorithm called SpectralRules that directly generates a compact pool of rules in Stage 1 with high diversity (of different levels of precision and recall). By design, SpectralRules generates a much smaller rule set (with no post-filtering needed) in Stage 1 with highly diversified rules, and benefits Stage 2 accordingly with final rule subsets of higher quality.

For finding the rule subset in Stage 2, they are usually evaluated by precision and recall~\cite{li2022adaptive}. Precision in our context is negatively correlated with the frequency that a user experiences an interruption for a transaction, and recall relates to the amount of fraudulent activities that can be prevented for a financial institution. Note that these two objectives naturally conflict with each other, and in practice one either seeks to maximize recall constrained on some precision threshold~\cite{li2022adaptive}, or combines the two metrics into a single generalized $F$ score that can balance the effects of the two objectives. $F$ score uses a parameter $\beta$ to control the contribution of the two inputs; $\beta$ is chosen such that recall is considered $\beta$ times as important as precision~\cite{vanrijsbergen1979information}. A large $\beta$ weighs recall higher than precision and a small $\beta$ is more biased towards precision.  
For fraud prevention problems in Fintech applications, typically a small $\beta$ is used to give more weights on precision, since usually the fraudulent ratio is low.

In practice, however, rule mining is an \emph{explorative} and \emph{iterative} process, and one rarely knows beforehand the best constraint threshold or $\beta$ for a particular problem. For example, one might have to experiment multiple precision thresholds and choose a rule subset according to subjective preferences or domain knowledge~\cite{li2022adaptive}. In such cases, multiple trials of the rule subset selection algorithm (Stage 2) are necessary, which is both time consuming and human labor intensive. Therefore, in this work we aim to find a set of Pareto-optimal solutions \emph{all at once} in a bi-objective space of precision and recall, i.e., rule subsets in which one objective cannot be improved without worsening the other one. A solution (rule subset) Pareto-dominates another one if both its precision and recall are no worse and at least one is strictly better. The image of this non-dominated solution set in the objective space is the so-called Pareto front. 

Finding Pareto-optimal rule subsets can be viewed as an intermediate stage between Stage 1 and Stage 2. It takes the initial pool of rules from Stage 1 and generates a set of non-dominated rule subsets. Finally, one can easily experiment different precision thresholds or $\beta$ on the (already generated) Pareto-optimal solution set and choose one as the final output for Stage 2. For example, if the goal is to maximize recall constrained on precision $\ge$ 0.9, one can simply filter all Pareto-optimal solutions whose precision is lower than 0.9 and choose the one with the highest recall, and therefore avoids the need to experiment multiple precision thresholds and greatly increases the flexibility for Stage 2.

The hypervolume indicator (HV) is one of the most widely used evaluation metric for the quality of a Pareto front~\cite{li2019quality}. It measures the ``size of the dominated space''~\cite{zitzler1999evolutionary}, and our goal for this paper is to find Pareto front of large HV. In this work, we mainly focus on a bi-objective space of precision and recall, and measure the dominated space referenced by (0, 0) (See Figure~\ref{fig:pf} for a visual example). Each solution on a Pareto front forms a rule subset.

This problem can be solved by standard evolutionary multi-objective optimization (EMO) algorithms, since bi-objective optimization is just a special case~\cite{li2015many, tanabe2017benchmarking}. Popular EMO algorithms include NSGA-II~\cite{deb2002fast}, MOEA/D~\cite{zhang2007moea}, etc. However, we empirically find EMO algorithms are not very efficient on our problem.

In this work, we propose a heuristic-based framework called PORS of finding Pareto-optimal solutions for rule subset selection problem. PORS iteratively ``expands'' the Pareto front by adding one rule to a solution (rule subset) on the front to see whether it expands the front (and thereby improves HV), and it terminates when no expansion of the current front can move the needle. Due to the combinatorial nature of this problem, it is computationally prohibitive to enumerate all rule subsets on the current Pareto front, and therefore we only select a set of \emph{representative} solutions on the current front as candidates for expansion to avoid exponential growth. We refer to this problem as solution selection on the front (SSF).

A closely related problem of SSF is the so-called subset selection of Pareto-optimal solutions, the goal of which is to reduce the number of solutions on a Pareto front to a user specified number while achieving some desired properties (e.g., a high HV approximation of the original Pareto front). Popular methods of Pareto subset selection include HV-SS~\cite{chen2021fast}, IGD/IGD+-SS~\cite{chen2021fast}, etc. Our motivation is different in that we are interested in finding a small set of promising solutions to expand on the current front so as to achieve a high HV when PORS \emph{terminates}. We can, nevertheless, employ existing algorithms in PORS framework.

Since finding a small set of the most promising solutions to expand on the current front is at the core of PORS framework, as one of our contributions, we propose a systematic categorization of solutions to this problem, map existing methods into different categories, and conduct a thorough empirical evaluation of methods in each category for SSF problem to study their effects on the quality of final Pareto front. If we think of the front as a ``curve'' in some space, we can broadly classify methods into two categories; uniform or non-uniform sampling on the ``curve.'' The space could be the objective space (such as bi-objective space of precision and recall considered in this paper) which is usually a coordinate system, or non-objective space (and therefore potentially non-coordinate system) where only distance metric of two rule subsets is defined (e.g., Jaccard distance between two rule subsets). Table~\ref{tbl:methods} summarizes existing methods and the methods proposed in this paper into each category.

\begin{table}[t]
\begin{center}
\begin{tabular}{c|c|l}
    \hline
    Sampling & Space & Method \\
    \hline
    \hline
    \multirow{3}{*}{Uniform} & \multirow{2}{*}{Objective} & \texttt{equi-spaced}~\cite{pereyra2013equispaced} \\
    & & \texttt{equi-dist} (this paper) \\
    \cline{2-3}
    & Non-objective & \texttt{equi-jaccard} (this paper) \\
    \hline
    \multirow{6}{*}{\shortstack{Non- \\ uniform}} & \multirow{4}{*}{Objective} & \texttt{hv-ss}~\cite{chen2021fast} \\
    & & \texttt{igd-ss}~\cite{chen2021fast} \\
    & & \texttt{igd+-ss}~\cite{chen2021fast} \\
    & & \texttt{hvc-ss} (this paper) \\
    & & \texttt{k-medoids-pr}~\cite{chen2021clustering} \\
    \cline{2-3}
    & Non-objective & \texttt{k-medoids-jaccard} (this paper) \\
    \hline
\end{tabular}
\end{center}
\caption{Categorization of SSF methods.}
\label{tbl:methods}
\end{table}

Although there are benchmark studies on subset selection of Pareto-optimal solutions in the context of EMO~\cite{shang2022benchmarking}, there is little research of finding Pareto-optimal rule subsets; neither EMO-based methods nor the greedy heuristic-based approaches considered in this paper. Due to the combinatorial nature of rule set selection, we believe our work provides a valuable piece of research in this area. Our empirical study on both public datasets and proprietary datasets from Alipay reveals that the proposed \texttt{hvc-ss} is the best SSF method, which achieves highest HV on almost all cases. 

Finally, we empirically demonstrate  the positive correlation between HV and the final objective for evaluating the quality of rule subsets through two case studies and one online experiment, and show that by optimizing HV, the quality of the refined rule subsets does improve accordingly.

In summary, we make the following contributions in this paper.
\begin{itemize}
    \item We propose a novel variant of the sequential covering algorithm called SpectralRules that directly generates a compact pool of rules with high diversity, and demonstrate on both public and proprietary datasets that SpectralRules improves HV over existing tree-based approaches~\cite{li2022adaptive} for Stage 1.
    \item We propose a heuristic-based framework called PORS for finding Pareto-optimal rule subsets in a bi-objective space as an intermediate stage between Stage 1 and Stage 2. At the core of this framework is the SSF problem.
    \item We provide a systematic categorization of the SSF problem and a thorough empirical evaluation of various SSF methods on both public and proprietary datasets. We find that the proposed \texttt{hvc-ss} is the best method for our problem.
    \item We present two case studies and one online experiment of using PORS in this paper to produce fraud prevention rule subsets for applications inside Alipay, and demonstrate the advantages of our methodology compared to existing work. 
    \item We deploy our method proposed in this paper in our internal Fanglue system~\cite{qian2023fanglue}, and release a repository of code that can fully reproduce our results on public datasets in this study to promote research efforts in this area~\cite{github_repo}.
\end{itemize}

The rest of this paper is organized as follows. Section~\ref{sec:related_work} presents related work. We present preliminaries in Section~\ref{sec:prelim}. Our approach is discussed in Section~\ref{sec:approach}. Experimental results are presented in Section~\ref{sec:exp} and we conclude the paper in Section~\ref{sec:conclusion}.

\section{Related Work}
\label{sec:related_work}

\subsection{Rule Set Mining}
There are generally two classes of approaches to generating rule sets. Tree based approaches, such as decision tree~\cite{breiman1984classification, quinlan1993c4} or random forests~\cite{breiman2001random}, extract rules for each path in the tree or tree ensemble. The other class of approaches focuses on direct rule induction to form rule sets. Popular methods of this category include sequential covering~\cite{molnar2020interpretable} (e.g., CN2~\cite{clark1989cn2} and RIPPER~\cite{cohen1995fast}) and association rule mining~\cite{agrawal1993mining}. Recently there is another line of research that focuses on the diversity of rule sets~\cite{lakkaraju2016interpretable, zhang2020diverse}. 

In practice, a two-stage framework of rule set mining is used in Fintech institutions such as Alipay~\cite{li2022adaptive}. A variant of tree ensemble is employed to generate an initial rule set (Stage 1), followed by a rule subset selection procedure that produces a refined rule subset according to some criteria (Stage 2). Stage 1 in~\cite{li2022adaptive} initially generates a large pool of rules, and therefore various filters are needed to reduce the number of rules for better efficiency and effectiveness. In addition, since Stage 2 directly produces a refined rule subset, multiple trials of Stage 2 are necessary as one rarely knows the selection criteria beforehand, which is both time consuming and human labor intensive.

\subsection{Evolutionary Multi-objective Optimization}
The evolutionary multi-objective optimization (EMO) algorithm is a popular choice for solving multi-objective optimization problems~\cite{li2015many, tanabe2017benchmarking}. EMO algorithms can be roughly classified into three categories; dominance-based approach, decomposition-based approach and indicator-based approach. Dominance based EMO algorithms commonly use Pareto dominance relationship and distance-based density estimation in the objective space for offspring generation and environmental selections. NSGA-II~\cite{deb2002fast} and SPEA2~\cite{zitzler2001spea2} are well-known and widely used dominance based EMO algorithms, which are believed to be very suitable for problems under 4 objectives~\cite{tanabe2017benchmarking}. Decomposition based approach decomposes a given multi-objective problem into single-objective sub-problems. MOEA/D is a representative decomposition based EMO algorithm~\cite{zhang2007moea}. Indicator based approaches employ a quantity indicator to evaluate or rank solutions. Popular approaches in this category include SMS-EMOA~\cite{beume2007sms}, HypE~\cite{bader2011hype}, etc. EMO algorithms are, however, typically inefficient on large problems. This paper employs a heuristic-based approach that achieves higher efficiency with better Pareto fronts.

\subsection{Subset Selection of Pareto-optimal Solutions}
Since the number of Pareto-optimal solutions can be very large for combinatorial optimization problems and infinity for continuous optimization problems, subset selection of Pareto-optimal solutions is usually an essential step. It is regarded as one of the most important topics in EMO domain, since it is involved in many phases of EMO algorithms~\cite{chen2021clustering, chen2021fast}. It is also studied in the context of discrete approximation to the Pareto front for continuous optimization problems~\cite{pereyra2013equispaced}. Some previous studies favor uniform sampling of the Pareto front~\cite{pereyra2013equispaced} while others favor non-uniform sampling~\cite{chen2021clustering, chen2021fast, guerreiro2005hypervolume}. 

In this work, we employ existing algorithms as SSF methods in our PORS framework and further extend this categorization into 4 categories as listed in Table~\ref{tbl:methods}, which is not revealed in existing literature. Under this view, we present 4 new methods (one for each category) and perform a thorough empirical evaluation on both public and proprietary datasets in the context of fraud prevention rule set mining for Fintech applications.

\section{Preliminaries}
\label{sec:prelim}

Let $\mathcal{D} = \{(\boldsymbol x_i, y_i)\}_{i=1}^N$ denote a dataset of size $N$, where $\boldsymbol x_i = (x_{i1}, ...,$ $x_{ip})$ is a feature vector with $p$ features for a transaction, and $y_i \in \{0, 1\}$ indicates whether the corresponding transaction is reported fraud or not. We use $\mathcal{P} = \{i|y_i = 1\}$ to denote the set of points with positive labels. We use $\delta = |\mathcal{P}| / N$ to denote the positive ratio of $\mathcal{D}$.

A rule $r$ has two parts; a conjunction of conditions and a prediction. A condition is of the form (feature, operator, value), e.g., \texttt{age >} 50. A rule makes a certain prediction when all conditions are satisfied for a given $\boldsymbol x_i$. In this case, we say the rule covers $\boldsymbol x_i$. Since typically the positive ratio $\delta$ for fraud in Fintech applications is very low, we only consider positive-class rules. Hence, $r(\boldsymbol x_i) = 1$ iff $r$ covers $\boldsymbol x_i$. A rule set $\mathcal{R}$ makes a positive prediction on $\boldsymbol x_i$ if there exists a rule $r \in \mathcal{R}$ such that $r(\boldsymbol x_i) = 1$. We use $\mathcal{D}(\mathcal{R}) = \{i|\mathcal{R}(\boldsymbol x_i) = 1\}$ to denote the set of data points covered by $\mathcal{R}$. 

\begin{figure}
\centering
\begin{tikzpicture}[thick, align=center, scale=0.65]

  \fill[black!20] (0,0) rectangle (5, 2.5);
  
  \fill[yellow!20] (0,0) rectangle (1, 4);
  \fill[yellow!20] (0,0) rectangle (2, 3.5);
  \fill[yellow!20] (0,0) rectangle (2.5, 2.5);
  \fill[yellow!20] (0,0) rectangle (4, 2);
  \fill[yellow!20] (0,0) rectangle (5, 1);
  
  \node[rotate=90, scale=1] at (-0.3, 2.5) {Precision};
  \node[scale=1] at (3, -0.3) {Recall};

  \node[rectangle, fill=green!30, scale=1.5, draw] at (1, 4) (a) {};
  \node[rectangle, fill=green!30, scale=1.5, draw] at (2, 3.5) (b) {};
  \node[rectangle, fill=green!30, scale=1.5, draw] at (2.5, 2.5) (c) {};
  \node[rectangle, fill=green!30, scale=1.5, draw] at (4, 2) (d) {};
  \node[rectangle, fill=green!30, scale=1.5, draw] at (5, 1) (e) {};
  \draw[thick, ->] (0,0) -- (6,0);
  \draw[thick, ->] (0,0) -- (0,5);
  \draw[very thick, -, color=red] (a) -- (1, 3.5) -- (b);
  \draw[very thick, -, color=red] (b) -- (2, 2.5) -- (c);
  \draw[very thick, -, color=red] (c) -- (2.5, 2) -- (d);
  \draw[very thick, -, color=red] (d) -- (4, 1) -- (e);

  \node[rectangle, fill=red!30, scale=1.5, draw] at (5, 2.5) (d) {};
  
  \node[scale=1] at (-0.1, -0.3) {$(0, 0)$};

\end{tikzpicture}
\caption{Illustration of Pareto dominance. A set of 5 non-dominated solutions (green square) constitutes the Pareto front (red line). Hypervolume of those 5 solutions is the size of the light yellow region. The hypervolume contribution of the solution (red square) to those 5 solutions is the size of the light grey region.}
\label{fig:pf}
\end{figure}
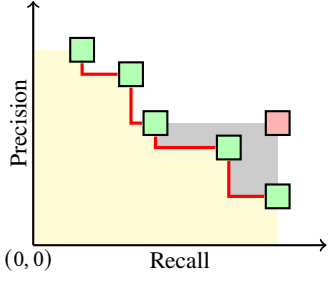

\begin{definition}[Precision]
We define the precision of a rule set $\mathcal{R}$ as the percentage that $\mathcal{R}$ correctly finds a fraud,
\begin{align}
\mbox{Prec}(\mathcal{R}) = |\mathcal{D}(\mathcal{R}) \cap \mathcal{P}|/|\mathcal{D}(\mathcal{R})|.
\end{align}
\end{definition}

\begin{definition}[Recall]
We define the recall of a rule set $\mathcal{R}$ as the ratio of the frauds covered by $\mathcal{R}$ among all fraudulent transactions,
\begin{align}
\mbox{Rec}(\mathcal{R}) = |\mathcal{D}(\mathcal{R}) \cap \mathcal{P}|/|\mathcal{P}|.
\end{align}
\end{definition}

\begin{definition}[$F_\beta$ score]
The general form of $F_\beta$ score for precision and recall is defined as,
\begin{align}
    F_\beta = (1 + \beta^2)\frac{\mbox{precision} \times \mbox{recall}}{(\beta^2 \times \mbox{precision}) + \mbox{recall}}.
\end{align}
\end{definition}

A large $\beta$ weighs recall higher than precision, and a small $\beta$ weighs precision higher than recall. For fraud prevention problems in Fintech applications, typically a small $\beta$ is used to give more weights on precision, since usually the positive (fraudulent) ratio is low.

For Stage 2, given an initial rule set $\mathcal{R}$, we mainly aim to find high-quality rule subsets of $\mathcal{R}$ in this bi-objective space of precision and recall.\footnote{Nevertheless, our methodology can be easily extended to metrics other than precision and recall.} To this end, we introduce the concept of Pareto dominance.

\begin{definition}[Pareto Dominance]
A rule subset $\mathcal{R}_1$ is Pareto-dominated by another rule subset $\mathcal{R}_2$ if $\mbox{Prec}(\mathcal{R}_1) \le \mbox{Prec}(\mathcal{R}_2) \wedge \mbox{Rec}(\mathcal{R}_1) \le \mbox{Rec}(\mathcal{R}_2)$ with at least one objective dimension (precision or recall) is strictly better. Under the context of Pareto dominance, we use rule subset and solution interchangeably. 
\end{definition}

\begin{definition}[Pareto Optimality]
A set of solutions (rule subsets) is called Pareto-optimal if no solution Pareto-dominates another. 
\end{definition}

\begin{definition}[Hypervolume Indicator]
Given a solution set $\mathcal{S}$ in a bi-objective space of precision and recall, the hypervolume indicator HV$(\mathcal{S})$ is the measure of region dominated by $\mathcal{S}$, referenced by (0, 0).
\end{definition}

Note that Pareto optimality only defines the situation where no solution can Pareto-dominate another \emph{within} a solution set. One can, nevertheless, compare two sets of solutions that are both Pareto-optimal (via HV).

\begin{definition}[Hypervolume Contribution]
Given two solution sets $\mathcal{S}$ and $\mathcal{T}$ in a bi-objective space of precision and recall, the hypervolume contribution of $\mathcal{T}$ to $\mathcal{S}$ referenced by (0, 0) is,
\begin{align}
    \mbox{HVC}(\mathcal{T}, \mathcal{S}) = \mbox{HV}(\mathcal{T} \cup \mathcal{S}) - \mbox{HV}(\mathcal{S} \setminus \mathcal{T}).
\end{align}
\end{definition}

Figure~\ref{fig:pf} illustrates a set of 5 non-dominated solutions (green square) in a precision-recall space, which constitutes the Pareto front (red line). Hypervolume of those 5 solutions is the size of the light yellow region. The hypervolume contribution of a new solution (red square) to those 5 solutions is the size of the light grey region.

\section{Our Approach}
\label{sec:approach}
In this section, we describe our methodology for finding high-quality rule subsets under the two-stage framework. We first introduce a novel algorithm called SpectralRules that generates a compact initial pool of rules $\mathcal{R}$ with high diversity for Stage 1 in Section~\ref{sec:sr}. Given a rule set $\mathcal{R}$, we present a heuristic-based framework called PORS that iteratively ``expands'' the Pareto front in Section~\ref{sec:framework} for finding high-quality rule subsets. We identify solution selection on the front (SSF) as the core problem of this framework, and provide a systematic categorization of this problem. In Section~\ref{sec:ssf}, we map existing work into different categories, and present new methods in each category.

\subsection{SpectralRules}
\label{sec:sr}

The quality of initial rule set from Stage 1 can largely impact the efficiency and efficacy of the rule subset selection. \cite{li2022adaptive} employs a variant of tree ensemble to extract the initial rule set, which may generate an unnecessarily large set of  ``homogeneous'' rules that look alike each other in terms of precision and recall (See Section~\ref{sec:sr_vs_tree_ens} for details). As noted in~~\cite{li2022adaptive}, various filters are needed to reduce the number of rules produced in this stage for better efficiency and effectiveness. Therefore, we introduce a novel variant of sequential covering algorithm called SpectralRules that promotes the diversity (in terms of precision and recall) of the initial pool of rules for Stage 1. By design, SpectralRules directly generates a much smaller rule set (with no post-filtering needed) in Stage 1 with high diversity, and benefits Stage 2 accordingly with final rule subsets of higher quality.

We first review the classic sequential covering algorithm (SCA). SCA is a ``separate-and-conquer'' procedure that repeatedly learns a single rule to form a rule set that covers the entire dataset rule by rule~\cite{lakkaraju2016interpretable}. Algorithm~\ref{algo:sc} summarizes the sequential covering algorithm. We start with empty rule set $\mathcal{R}$ and whole dataset $\mathcal{D}$ (Line 2-3). We iteratively learn a rule $r$ using standard rule induction procedure that employs $F_\beta$ as the evaluation metric to grow a rule up to a pre-specified length $len$, and then remove all points covered by $r$ from the remaining dataset, until we find $n$ rules (Line 4-7).

\begin{algorithm}[t]
\caption{Sequential Covering}
\label{algo:sc}
\begin{algorithmic}[1]
	\Procedure{SequentialCovering}{$\mathcal{D}$, $n$, $len$, $\beta$}
		\State $\mathcal{R} \leftarrow \emptyset$
		\State $\mathcal{D}' \leftarrow \mathcal{D}$
		\For{$i = 1$ to $n$}
			\State $r \leftarrow \textsc{RuleInduction}(len, \beta, \mathcal{D}')$
			\State $\mathcal{D}' \leftarrow \mathcal{D}' \setminus \{\boldsymbol x~|~r~\mbox{covers}~\boldsymbol x,  \boldsymbol x\in\mathcal{D}'\}$
			\State $\mathcal{R} \leftarrow \mathcal{R} \cup \{r\}$
		\EndFor
		\Return $\mathcal{R}$
	\EndProcedure
\end{algorithmic}
\end{algorithm}

\begin{algorithm}[t]
\caption{SpectralRules}
\label{algo:sr}
\begin{algorithmic}[1]
	\Procedure{SpectralRules}{$\mathcal{D}$, $n$, $len$, $\mathcal{B}$}
		\State $\mathcal{R} \leftarrow \emptyset$
		\For{$\beta \in \mathcal{B}$}
			\State $\mathcal{R} \leftarrow \mathcal{R} \cup \textsc{SequentialCovering}(\mathcal{D}, \lc\frac{n}{|\mathcal{B}|}\rc, len, \beta)$
		\EndFor
		\Return $\mathcal{R}$
	\EndProcedure
\end{algorithmic}
\end{algorithm}

A very small $\beta$ (such as 0.1) can be employed to focus on mining high precision rules. However, recall of the rules produced by SCA might be quite limited. As a result, we may end up with many rules of high precision and low recall, leading to a suboptimal rule set. To this end, we propose SpectralRules to encourage the diversity among rules (i.e., rules with different levels of precision and recall). This is achieved through multiple trials of Algorithm~\ref{algo:sc} with $\beta$ ranging from small number (such as 0.01) to large number (such as 1). 

Algorithm~\ref{algo:sr} summarizes the SpectralRules procedure. We are given a set $\mathcal{B}$ with multiple $\beta$'s. For each $\beta\in\mathcal{B}$, we run Algorithm~\ref{algo:sc} with this specific $\beta$ to generate at most $\lc\frac{n}{|\mathcal{B}|}\rc$ rules (Line 3-4). The resulting rule set $\mathcal{R}$ is composed of rules with different levels of precision and recall. In this paper, we fix $\mathcal{B}=\{0.01, 0.02, 0.04, 0.06, 0.08, 0.10, 0.20, 0.40, 0.60, 0.80\}$. As we will see in our experiments, SpectralRules usually leads to Pareto fronts with higher HV (Section~\ref{sec:main_results}), and benefits Stage 2 accordingly with final rule subsets of higher quality (Section~\ref{sec:case_study}).

\subsection{The PORS Framework}
\label{sec:framework}
Given an initial rule set $\mathcal{R}$ from Stage 1, PORS iteratively ``expands'' the current Pareto front by adding one rule to a solution (rule subset) on the front to form a new set of solutions. Due to the combinatorial nature of this problem, \naive implementation takes $O(|\mathcal{R}|^n + c)$ time for the $n$-th iteration (where $c$ denotes the complexity of computing Pareto front on the newly formed set of solutions), it is computationally prohibitive to enumerate all rule subsets on the current Pareto front, and therefore we aim to select a set of \emph{representative} solutions on the current front as candidates for expansion. We refer to this problem as solution selection on the front (\textsc{SSF}).

Algorithm~\ref{algo:framework} summarizes the proposed PORS framework. The framework starts with singleton sets for each rule in $\mathcal{R}$, and then makes them a Pareto front via \textsc{makePF} procedure by retaining only non-dominated solutions (Line 1). PORS applies some \textsc{SSF} method to sample $k$ solutions on the current front in order to avoid exponential growth (Line 5). Those $k$ solutions are then expanded by adding one rule from the initial pool of rules $\mathcal{R}$ to form new solutions (Line 6-10). The expanded solutions are then put together with the solutions from last front to form a new Pareto front by retaining only non-dominated solutions (Line 11). The framework terminates when the current Pareto front does not change after expansion (Line 12), otherwise the current front is ready for the next iteration (Line 13).

The time complexity per iteration of PORS now becomes $O(k|\mathcal{R}| + c)$, which is much more computationally affordable compared to naive implementation. In addition, we empirically found all SSF methods are very efficient for different values of $k$ (Appendix~\ref{sec:impact_k}). 
It is obvious that the performance of PORS is largely affected by the efficacy of the SSF method, and therefore we discuss different candidate SSF methods in the next section.

\begin{algorithm}[t]
\caption{The Pareto Optimal Rule Subset Selection Framework}
\label{algo:framework}
\begin{algorithmic}[1]
    \State $\mathcal{F} \leftarrow \textsc{makePF}(\{\{r\} | r \in \mathcal{R}\})$
    \State $\texttt{converged} \leftarrow \texttt{false}$
    \While{\texttt{not} $\texttt{converged}$}
        \State $\texttt{converged} \leftarrow \texttt{true}$
        \State $\mathcal{F}_0 \leftarrow \textsc{SSF}(\mathcal{F}, k)$
        \State $\mathcal{F}' \leftarrow \emptyset$
        \For{$S \in \mathcal{F}_0$}
            \For{$r \in \mathcal{R}$}
                \State $S' \leftarrow S \cup \{r\}$
                \State $\mathcal{F}' \leftarrow \mathcal{F}' \cup \{S'\}$
            \EndFor
        \EndFor
        \State $\mathcal{F}' \leftarrow \textsc{makePF}(\mathcal{F}' \cup \mathcal{F})$
        \State $\texttt{converged} \leftarrow \mathcal{F}' == \mathcal{F}$
        \State $\mathcal{F} \leftarrow \mathcal{F}'$
    \EndWhile
    \Return $\mathcal{F}$
\end{algorithmic}
\end{algorithm}

\subsection{SSF Methods}
\label{sec:ssf}
We broadly classify SSF methods into two categories; uniform or non-uniform sampling on the ``curve'' of a Pareto front. Under each category, we identify two sub-categories depending on whether the ``distance measure'' is in the original objective space or not. 

\subsubsection{Uniform Sampling}

We first discuss methods by uniform sampling in the objective space. We consider two methods as follows.
\begin{itemize}
    \item \texttt{equi-spaced}~\cite{pereyra2013equispaced}. \cite{pereyra2013equispaced} considers uniform sampling of the Pareto front in the objective space on continuous problems. In our problem, it is equivalent to uniform sampling of $k$ solutions by walking along a Pareto front curve with roughly the same Manhattan distance for two consecutive samples.
    \item \texttt{equi-dist}. Following the same idea, we propose uniform sampling of $k$ solutions on a Pareto front with roughly the same Euclidean distance for two consecutive samples.
\end{itemize}

The main difference between these two methods is whether the distance is measured on curve (Manhattan) or not (Euclidean).

We now describe a method called \texttt{equi-jaccard} that performs uniform sampling in non-objective space. We employ Jaccard distance to measure the dissimilarity between the coverage of two rule subsets. Formally, for two rule subsets $\mathcal{R}_1$ and $\mathcal{R}_2$, their Jaccard distance is defined as $\texttt{Jaccard}(\mathcal{R}_1, \mathcal{R}_2) = 1 - \frac{|\mathcal{D}(\mathcal{R}_1) \cap \mathcal{D}(\mathcal{R}_2)|}{|\mathcal{D}(\mathcal{R}_1) \cup \mathcal{D}(\mathcal{R}_2)|}$.

Since this is a non-coordinate system, we think of all solutions on a Pareto front as a clique where edges between two solutions are weighed by their Jaccard distance. Now we can generate a traversal of all solutions on this graph to form a ``curve'' in this space, and then apply the same idea of \texttt{equi-spaced} so that Jaccard distance is roughly the same for two consecutive samples. There are many ways to perform a traversal. We can perform depth-first search on this graph to minimize or maximize the Jaccard distance of the overall path. We can also search for minimum sum of Jaccard distance of the traversal using the travelling salesperson (TSP) algorithm.\footnote{TSP generates a loop with minimized overall distance. We cut the edge with the largest Jaccard distance on the loop to form a traversal path.} During our early investigation, we found that different methods lead to the final Pareto fronts of similar HV. Therefore, we only report results of \texttt{equi-jaccard} on a traversal generated by TSP algorithm.

\subsubsection{Non-uniform Sampling}
We consider the following methods of non-uniform sampling in the objective space.
\begin{itemize}
    \item \texttt{hv-ss}~\cite{chen2021fast}. \texttt{hv-ss} aims to select $k$ solutions so that their hypervolume is maximized.
    \item \texttt{igd/igd+-ss}~\cite{chen2021fast}. \texttt{igd/igd+-ss} aims to select $k$ solutions so that their Inverted Generational Distance (IGD) or Inverted Generational Distance plus (IGD+) is minimized.
    \item \texttt{hvc-ss}. We propose in this work to select $k$ solutions so that their HVC to the last Pareto front is maximized. This is achieved using a greedy forward stepwise approach.
    \item \texttt{k-medoids-pr}~\cite{chen2021clustering} (or \texttt{k-med.-pr} for short). This is the standard $k$-medoids clustering performed in the objective space, with distance measured by Euclidean distance.
\end{itemize}

We also consider $k$-medoids clustering performed on Jaccard distance measure, and introduce a non-uniform sampling in non-objective space method called \texttt{k-medoids-jaccard} (or \texttt{k-med.-j.} for short).

Some of the aforementioned methods are studied in previous research in the context of EMO on continuous problems~\cite{shang2022benchmarking}, but there is little research of all those 9 methods (covering 4 categories of the SSF problem) for combinatorial problems of finding Pareto-optimal rule subsets considered in this paper. To this end, we provide a thorough evaluation of those 9 methods in Section~\ref{sec:exp}.

\section{Experiments}
\label{sec:exp}

We report our results on both public datasets and proprietary datasets from Alipay. We mainly compare our approach with~\cite{li2022adaptive} under the same two-stage framework for the same application domain. Note that~\cite{li2022adaptive} directly outputs a refined rule subset according to a given precision threshold (measured using the final objective, i.e., the recall) while we aim to generate a set of non-dominated solutions (to simply the rule subset selection in Stage 2) and therefore an intermediate metric HV is introduced. The experiments in this section are designed with care to answer the following research questions.
\begin{itemize}
    \item[Q1] Whether SpectralRules and PORS improve HV?
    \item[Q2] Which SSF method achieves the highest HV in PORS?
    \item[Q3] Whether large HV suggests high quality (final objective) of the refined rule subsets? 
\end{itemize}

We release a repository of code that can fully reproduce the results (including tables and figures) in this section on public datasets~\cite{github_repo}.

\begin{table}[t]
    \centering
    \begin{tabular}{c|c|c|c}
    \hline
    Dataset & Size & Attributes & Pos. Ratio \\
    \hline
    \hline
    Default & 30,000 & 24 & 22.12\% \\
    Credit & 150,000 & 11 & 6.68\% \\
    Fraud & 284,807 & 31 & 0.17\% \\
    Bank & 45,211 & 17 &  11.70\% \\
    \hline
    A1 & 6,609,266 & 731 & $\delta_1$ \\
    A2 & 990,798 & 210 & $\delta_2$ \\
    A3 & 9,837,541 & 176 & $\delta_3$ \\
    \hline
    \end{tabular}
    \caption{Datasets.}
    \label{tbl:datasets}
\end{table}

{\small
\begin{table*}[t]
\begin{center}
\begin{tabular}{c|c|c||c|c|c|c||c|c|c}
    \hline
    \multicolumn{2}{c|}{Type} & Method & Default & Credit & Fraud & Bank & A1 & A2 & A3\\
    \hline
    \hline
    \multicolumn{2}{c|}{EMO} & \texttt{NSGA-II}~\cite{deb2002fast} & 0.524$\pm$0.013  & 0.322$\pm$0.023 & 0.786$\pm$0.027 & 0.533$\pm$0.018 & 0.296$\pm$0.010 & 0.169$\pm$0.006 & 0.032$\pm$0.006 \\
    \hline
    \multirow{3}{*}{unif.} & \multirow{2}{*}{obj.} & \texttt{equi-spaced}~\cite{pereyra2013equispaced} & 0.515$\pm$0.008 & 0.323$\pm$0.021 & 0.783$\pm$0.020 & 0.517$\pm$0.020 & 0.299$\pm$0.011 & 0.172$\pm$0.006 & 0.033$\pm$0.006 \\
    \cline{3-10}
    &  & \texttt{equi-dist} & 0.517$\pm$0.010  & 0.324$\pm$0.022 & 0.786$\pm$0.021 & 0.528$\pm$0.014 & 0.298$\pm$0.010 & 0.173$\pm$0.006 & 0.034$\pm$0.006 \\
    \cline{2-10}
    & non-obj. & \texttt{equi-jaccard} & 0.509$\pm$0.007  & 0.308$\pm$0.018 & 0.789$\pm$0.032 & 0.516$\pm$0.019 & 0.296$\pm$0.012 & 0.166$\pm$0.009 & 0.033$\pm$0.006 \\
    \hline
    \multirow{5}{*}{\shortstack{non- \\ unif.}} & \multirow{4}{*}{obj.} & \texttt{hv-ss}~\cite{chen2021fast} & 0.519$\pm$0.009 & 0.325$\pm$0.022 & 0.783$\pm$0.037 & 0.530$\pm$0.021 & 0.298$\pm$0.007 & 0.165$\pm$0.007 & 0.034$\pm$0.005 \\
    \cline{3-10}
     &  & \texttt{igd-ss}~\cite{chen2021fast} & 0.496$\pm$0.012  & 0.297$\pm$0.023 & 0.790$\pm$0.038 & 0.499$\pm$0.021 & 0.297$\pm$0.012 & 0.163$\pm$0.006 & 0.031$\pm$0.007 \\
    \cline{3-10}
     &  & \texttt{igd+-ss}~\cite{chen2021fast} & 0.491$\pm$0.010  & 0.281$\pm$0.018 & 0.789$\pm$0.031 & 0.480$\pm$0.025 & 0.289$\pm$0.011 & 0.147$\pm$0.008 & 0.027$\pm$0.004  \\
    \cline{3-10}
     &  & \texttt{hvc-ss} & \textbf{0.524$\pm$0.011}  & \textbf{0.325$\pm$0.022} & \textbf{0.798$\pm$0.040} & \textbf{0.533$\pm$0.015} & \textbf{0.301$\pm$0.012} & \textbf{0.176$\pm$0.006} & \textbf{0.036$\pm$0.006} \\
    \cline{3-10}
     &  & \texttt{k-medoids-pr}~\cite{chen2021clustering} & 0.501$\pm$0.013  & 0.303$\pm$0.025 & 0.785$\pm$0.027 & 0.511$\pm$0.020 & 0.296$\pm$0.014 & 0.170$\pm$0.005 & 0.031$\pm$0.007 \\
    \cline{2-10}
     & non-obj. & \texttt{k-medoids-jaccard} & 0.472$\pm$0.016  & 0.258$\pm$0.035 & 0.792$\pm$0.034 & 0.479$\pm$0.026 & 0.284$\pm$0.012 & 0.142$\pm$0.006  & 0.024$\pm$0.010 \\
    \hline
\end{tabular}
\end{center}
\caption{The hypervolume (HV) performance (mean$\pm$std) of the Pareto front for different methods on test set of each problem. Higher HV is better and best method for each dataset is marked in \textbf{bold}. TreeEns~\cite{li2022adaptive} is employed in Stage 1 to produce the initial 500 rules.}
\label{tbl:results_trees}
\end{table*}
}

{\small
\begin{table*}[t]
\begin{center}
\begin{tabular}{c|c|c||c|c|c|c||c|c|c}
    \hline
    \multicolumn{2}{c|}{Type} & Method & Default & Credit & Fraud & Bank & A1 & A2 & A3\\
    \hline
    \hline
    \multicolumn{2}{c|}{EMO} & \texttt{NSGA-II}~\cite{deb2002fast} & 0.526$\pm$0.013  & 0.350$\pm$0.011 & 0.789$\pm$0.036 & 0.585$\pm$0.011 & 0.313$\pm$0.004 & 0.188$\pm$0.005 & 0.033$\pm$0.004 \\
    \hline
    \multirow{3}{*}{unif.} & \multirow{2}{*}{obj.} & \texttt{equi-spaced}~\cite{pereyra2013equispaced} & 0.534$\pm$0.012  & 0.350$\pm$0.014 & 0.796$\pm$0.037 & 0.570$\pm$0.009 & 0.313$\pm$0.006 &  0.183$\pm$0.007 & 0.035$\pm$0.006 \\
    \cline{3-10}
    &  & \texttt{equi-dist} & 0.535$\pm$0.011  & 0.352$\pm$0.015 & 0.805$\pm$0.025 & 0.575$\pm$0.009 & 0.315$\pm$0.005 & 0.180$\pm$0.008 & 0.034$\pm$0.006 \\
    \cline{2-10}
    & non-obj. & \texttt{equi-jaccard} & 0.535$\pm$0.014 & 0.357$\pm$0.014 & 0.803$\pm$0.028 & 0.577$\pm$0.013 & 0.314$\pm$0.009 & 0.176$\pm$0.008 & 0.034$\pm$0.007 \\
    \hline
    \multirow{5}{*}{\shortstack{non- \\ unif.}}  & \multirow{4}{*}{obj.} & \texttt{hv-ss}~\cite{chen2021fast} & 0.534$\pm$0.011  & 0.357$\pm$0.016 & 0.801$\pm$0.033 & 0.582$\pm$0.008 & 0.318$\pm$0.004 & 0.184$\pm$0.006 & 0.035$\pm$0.005 \\
    \cline{3-10}
     &  & \texttt{igd-ss}~\cite{chen2021fast} & 0.516$\pm$0.015  & 0.346$\pm$0.016 & 0.798$\pm$0.045 & 0.560$\pm$0.012 & 0.317$\pm$0.006 & 0.185$\pm$0.006 & 0.033$\pm$0.007 \\
    \cline{3-10}
     &  & \texttt{igd+-ss}~\cite{chen2021fast} & 0.515$\pm$0.016  & 0.338$\pm$0.012 & 0.793$\pm$0.024 & 0.553$\pm$0.017 & 0.308$\pm$0.005 & 0.160$\pm$0.007 & 0.029$\pm$0.004 \\
    \cline{3-10}
     &  & \texttt{hvc-ss} & \textbf{0.540$\pm$0.012} & \textbf{0.359$\pm$0.013} & \textbf{0.808$\pm$0.030} & \textbf{0.593$\pm$0.012} & \textbf{0.319$\pm$0.006} & \textbf{0.192$\pm$0.007} & \textbf{0.036$\pm$0.005} \\
    \cline{3-10}
     &  & \texttt{k-medoids-pr}~\cite{chen2021clustering} & 0.515$\pm$0.013 & 0.343$\pm$0.014 & 0.801$\pm$0.033 & 0.563$\pm$0.009 & 0.317$\pm$0.004 & 0.184$\pm$0.005 & 0.033$\pm$0.006  \\
    \cline{2-10}
     & non-obj. & \texttt{k-medoids-jaccard} & 0.507$\pm$0.016  & 0.333$\pm$0.011 & 0.791$\pm$0.030 & 0.521$\pm$0.014 &  0.301$\pm$0.011 & 0.151$\pm$0.006 & 0.028$\pm$0.008 \\
    \hline
\end{tabular}
\end{center}
\caption{The hypervolume (HV) performance (mean$\pm$std) of the Pareto front for different methods on test set of each problem. Higher HV is better and best method for each dataset is marked in \textbf{bold}. SpectralRules is employed in Stage 1 to produce the initial 500 rules.}
\label{tbl:results_sr}
\end{table*}
}

\subsection{Datasets}
\label{sec:dataset}

Table~\ref{tbl:datasets} summarizes the datasets used in our experiments. We use 4 public Fintech-related datasets in this paper. ``Default'' and ``Bank'' are from the UCI repository.\footnote{\url{http://archive.ics.uci.edu/ml/}} ``Credit'' is a binary classification problem that predicts the probability that somebody will experience financial distress in the next two years.\footnote{\url{https://www.kaggle.com/c/GiveMeSomeCredit/data}} ``Fraud'' is a Kaggle competition problem that aims to recognize fraudulent credit card transactions.\footnote{\url{https://www.kaggle.com/mlg-ulb/creditcardfraud}} 

``A1'', ``A2'' and ``A3'' are three proprietary fraud prevention problems inside Alipay, representing three different fraudulent activities such as fraudulent payment transactions, identity theft, etc. Those datasets are large in scale and we do not reveal the positive ratio of each proprietary dataset by using $\delta$ to symbolize the actual number.

\subsection{Pareto-optimal Rule Subsets (Q1 \& Q2)}
\label{sec:main_results}
We first conduct a thorough evaluation of finding Pareto-optimal rule subsets. We consider the classic EMO algorithm NSGA-II~\cite{deb2002fast} which is believed to perform well for bi-objective optimization problems~\cite{tanabe2017benchmarking}.\footnote{We use pyMultiobjective package at \url{https://github.com/Valdecy/pyMultiobjective}. We also experimented HypE~\cite{bader2011hype} and MOEA/D~\cite{zhang2007moea} during our early investigation, but we found NSGA-II is more efficient and achieves higher HV.}  Following standard practice~\cite{chen2021fast}, we equip NSGA-II with an unbounded external archive (UEA) to store all non-dominated solutions examined during the execution of the algorithm. We found that NSGA-II with UEA consistently achieves higher performance. We set the parameter $mutation\_rate$ to 0.02 and $generations$ to 1000 while using default values for other parameters. We observe this parameter setup often achieves the best results for NSGA-II. For PORS framework, we consider all SSF methods listed in Section~\ref{sec:ssf} to implement the framework. We use $k=10$ as the number of samples for SSF, which is the only parameter to set for SSF methods.

We consider two candidates for generating the initial rule set in Stage 1. We follow the same approach in~\cite{li2022adaptive} that uses a variant of tree ensemble (called TreeEns), and compare that with SpectralRules for generating the initial set of rules. For both methods, maximum rule length is fixed to 6 and we generate (up to) 500 rules from each method to form the initial rule set $\mathcal{R}$ (same setting as in~\cite{li2022adaptive}).\footnote{For TreeEns, we first generate a large pool of rules and then apply the filter provided in~\cite{li2022adaptive} to reduce the size of rule set to pre-specified number as the output of Stage 1.}

We use hypervolume indicator (HV) to measure the quality of Pareto-optimal rule subsets. For all experiments, we randomly partition the dataset into training (60\% of the data), validation (20\% of the data) and test sets (20\% of the data). We generate the initial rule set in Stage 1 using training set. We then apply our 9 PORS algorithms and NSGA-II on the training set, select the best Pareto front with highest HV on the validation set and collect its HV on test set. All experiments are repeated 5 times and we report mean and standard deviation of the evaluation results on test sets. 

Table~\ref{tbl:results_trees} and~\ref{tbl:results_sr} shows the HV results on test set for each dataset when the initial rule set is generated by TreeEns and SpectralRules, respectively. We first observe that PORS with \texttt{hvc-ss} as its SSF method consistently achieves the highest HV for both cases. Therefore, we recommend implementing PORS framework with \texttt{hvc-ss} as its SSF method, and we now refer to this particular combination as the PORS algorithm. In addition, we also observe that employing SpectralRules in Stage 1 leads to higher HV of the final Pareto front by comparing the best values in Table~\ref{tbl:results_trees} and~\ref{tbl:results_sr}.

\textbf{Intuitive explanation of why \texttt{hvc-ss} performs the best}. PORS framework can be viewed as a greedy forward stagewise algorithm, with maximizing HV as its objective, by iteratively expanding Pareto fronts. Among all SSF methods, \texttt{hvc-ss} achieves the largest HV increase (by definition) for each iteration (Pareto front expansion), and therefore it should work well inside PORS framework. Because of the greedy nature of the framework, other SSF methods might also work well. Our empirical study complements the picture and shows \texttt{hvc-ss} indeed achieves the best results on most cases.

\subsection{Discussion}
In this section, we provide more experimental results to shed light on properties of different combinations of methods. 

\subsubsection{Comparison of SpectralRules and TreeEns (Stage 1)}
\label{sec:sr_vs_tree_ens}
It is apparent that the quality of the initial rule set $\mathcal{R}$ from Stage 1  can significantly impact the final Pareto front; it is impossible to find high-quality rule subsets when some key rules are not present in $\mathcal{R}$ in the first place. Section~\ref{sec:main_results} demonstrates that SpectralRules outperforms TreeEns when the size of $\mathcal{R}$ is set to 500. In this section, we provide further investigation on the effects of SpectralRules and TreeEns when the size of $\mathcal{R}$ varies. In this experiment, we use PORS algorithm as our method to find Pareto-optimal rule subsets.

\begin{figure}[t]
\begin{center}
\includegraphics[width=0.4\textwidth]{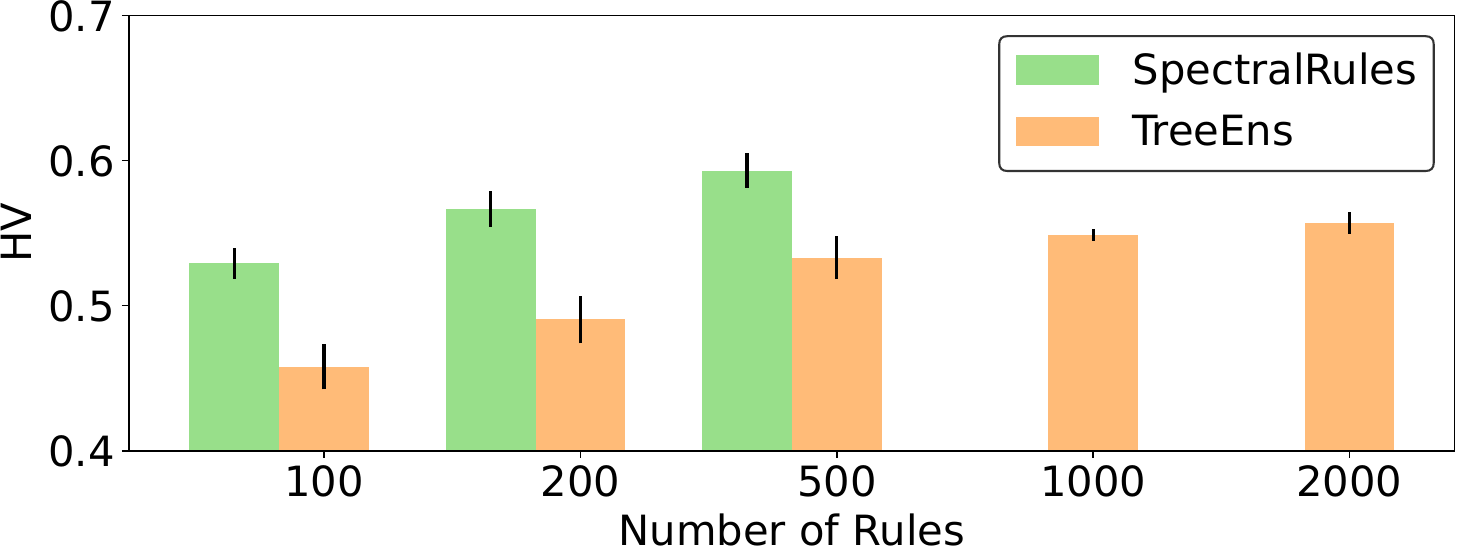}
\end{center}
\caption{HV of PORS algorithm on Bank dataset when using SpectralRules or TreeEns to produce rule set in Stage 1.}
\label{fig:sr_vs_tree}
\end{figure}

Figure~\ref{fig:sr_vs_tree} illustrates the HV performance on Bank dataset when the size of $\mathcal{R}$ increases. Results are aggregated over 5 trials of random partitioning of the dataset. We observe that SpectralRules outperforms TreeEns for all $|\mathcal{R}| \le 500$. 
SpectralRules cannot produce a rule set large enough to reach at size of 1000 since all positive points are already covered.
We also observe that even if we give more budget on the number of rules to TreeEns, it cannot catch up with the level of HV achieved by SpectralRules. Similar patterns are observed on other datasets.

There exist some key differences between SpectralRules and TreeEns. TreeEns extracts rules from a tree ensemble, and therefore we expect that there are many ``homogeneous'' rules (in terms of the distribution of precision and recall) that look alike each other. This not only hurts the efficiency of the subsequent subset selection (for there is now a large initial set to select from) but the efficacy of Stage 2 also takes a hit. As noted in~~\cite{li2022adaptive}, various filters are needed for TreeEns to reduce the number of rules produced in this stage for better efficiency and effectiveness. On the other hand, SpectralRules employs direct rule induction and focuses on promoting the diversity of $\mathcal{R}$, leading to a more compact rule set of higher quality.

On Bank dataset, for $|\mathcal{R}| = 500$, we observe the mean and standard deviation for precision (recall) of TreeEns and SpectralRules is $0.5461\pm0.1604$ ($0.0225\pm0.0330$) and $0.6376\pm0.1823$ ($0.0381\pm0.0699$), respectively. We see that not only the mean of precision (recall) from SpectralRules is higher, its standard deviation is larger as well; SpectralRules does generate rule set with high diversity. Combined with results in Section~\ref{sec:main_results}, we conclude that high diversity in Stage 1 improves HV. Hence, for the rest of this paper, we use SpectralRules for Stage 1 by default unless otherwise mentioned. 

\begin{figure}[t]
\begin{center}
\includegraphics[width=0.4\textwidth]{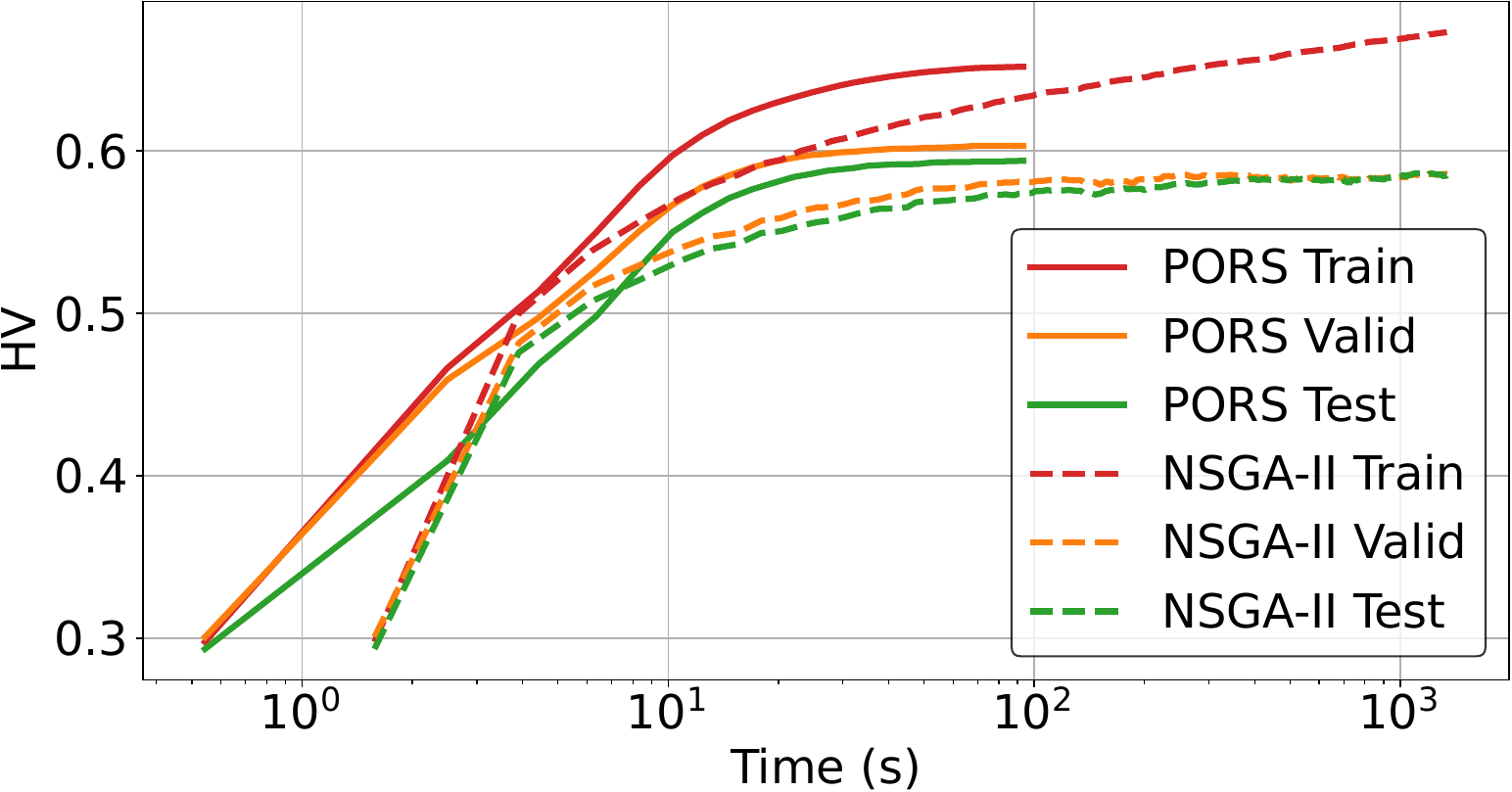}
\end{center}
\caption{Efficiency of NSGA-II vs. PORS on Bank dataset.}
\label{fig:emo_comp}
\end{figure}

\subsubsection{Efficiency Evaluation for EMO and PORS Algorithm}
\label{sec:emo_vs_pors}
From Section~\ref{sec:main_results} we observe that NSGA-II as an EMO algorithm is very competitive to PORS algorithm (as a heuristic-based algorithm). In this section, we conduct efficiency evaluation for these two methods. We run both algorithms on a standard server with 16GB memory using 4 threads.

Figure~\ref{fig:emo_comp} shows how HV changes on train, validation and test sets of Bank dataset over wall clock time for NSGA-II and PORS algorithm. We can see that PORS algorithm quickly converges while NSGA-II is less efficient. Similar patterns are observed on other datasets. In particular, on our largest dataset ``A3,'' PORS only takes 146 seconds to finish while NSGA-II needs 1610 seconds on the same server with 16GB using 4 threads. This suggests a heuristic-based algorithm can be highly efficient.

\subsection{Case Study (Q3)}
\label{sec:case_study}

In this section, we provide two case studies within Alipay where the requirement of the final rule subset of Stage 2 differs, and demonstrate how finding Pareto-optimal rule sets as an intermediate stage between Stage 1 and Stage 2 can benefit the eventual result. 

To study the relationship of HV (as an intermediate evaluation metric) and final objective, we consider three implementations of PORS framework in this section; \texttt{hvc-ss} for high HV, \texttt{equi-spaced} for medium HV, and \texttt{k-med.-j.} for low HV.

\subsubsection{Confidence-Constrained Rule Set Learning}
The first case study focuses on applications that prioritize recall over precision. We compare PORS with confidence-constrained rule set learning (CRSL)~\cite{li2022adaptive}. Note that CRSL only works on TreeEns as it needs a large candidate set to select from, and we compare it with PORS algorithms (with different SSF implementations) using SpectralRules for Stage 1 (other settings are the same with Section~\ref{sec:main_results}).

{\footnotesize
\begin{table}[t]
\begin{center}
\begin{tabular}{cccccc}
\hline
\multirow{3}{*}{Dataset} & \multirow{3}{*}{Prec} & \multirow{3}{*}{CRSL~\cite{li2022adaptive}} & \multicolumn{3}{c}{PORS} \\
& & & \texttt{hvc-ss} & \texttt{equi-spaced} & \texttt{k-med.-j.} \\
& & & (High HV) & (Med. HV) &  (Low HV) \\
\hline\hline
\multirow{3}{*}{Default} & 0.3 & 0.765$\pm$0.025 & \textbf{0.854$\pm$0.010} & 0.853$\pm$0.013 & 0.835$\pm$0.014 \\
& 0.5 & 0.519$\pm$0.012 & 0.557$\pm$0.020 & \textbf{0.564$\pm$0.005} & 0.543$\pm$0.017 \\
& 0.7 & 0.238$\pm$0.038 & 0.260$\pm$0.040 & \textbf{0.262$\pm$0.052} & 0.257$\pm$0.045 \\
\hline
\multirow{3}{*}{Credit} & 0.3 & 0.555$\pm$0.030  & \textbf{0.584$\pm$0.013}  & 0.581$\pm$0.015 & 0.538$\pm$0.012 \\
& 0.5 & 0.269$\pm$0.059 & \textbf{0.289$\pm$0.017} & 0.279$\pm$0.037 & 0.273$\pm$0.030 \\
& 0.7 & 0.038$\pm$0.021 & \textbf{0.041$\pm$0.013} & 0.025$\pm$0.023 & 0.024$\pm$0.023 \\
\hline
\multirow{3}{*}{Fraud} & 0.3 & 0.840$\pm$0.038 & \textbf{0.851$\pm$0.045} & 0.847$\pm$0.035 & 0.836$\pm$0.037 \\
& 0.5 & 0.816$\pm$0.024 & 0.822$\pm$0.035 & 0.822$\pm$0.038 & \textbf{0.822$\pm$0.024} \\
& 0.7 & 0.812$\pm$0.032 & 0.814$\pm$0.021 & \textbf{0.818$\pm$0.033} & 0.808$\pm$0.034 \\
\hline
\multirow{3}{*}{Bank} & 0.3 & 0.907$\pm$0.015 & \textbf{0.951$\pm$0.007} & 0.945$\pm$0.009 & 0.813$\pm$0.034 \\
& 0.5 & 0.630$\pm$0.075 & \textbf{0.781$\pm$0.018} & 0.764$\pm$0.034 & 0.679$\pm$0.050 \\
& 0.7 & \textbf{0.337$\pm$0.044} & 0.264$\pm$0.044 & 0.243$\pm$0.038 & 0.253$\pm$0.055 \\
\hline\hline
\multirow{3}{*}{A1} & 5$\delta_1$ & 0.789$\pm$0.023 & \textbf{0.824$\pm$0.013} & 0.794$\pm$0.016 & 0.775$\pm$0.021 \\
& 10$\delta_1$ & 0.624$\pm$0.016 & \textbf{0.657$\pm$0.011}  & 0.640$\pm$0.012 & 0.614$\pm$0.018 \\
& 20$\delta_1$ & 0.499$\pm$0.008 & \textbf{0.510$\pm$0.009}  & 0.501$\pm$0.006 & 0.487$\pm$0.009 \\
\hline
\multirow{3}{*}{A2} & 5$\delta_2$ & 0.828$\pm$0.019 & \textbf{0.831$\pm$0.007}  & 0.767$\pm$0.012 & 0.726$\pm$0.009 \\
& 10$\delta_2$ & 0.483$\pm$0.072  & \textbf{0.593$\pm$0.008} & 0.578$\pm$0.008  & 0.545$\pm$0.009 \\
& 20$\delta_2$ & 0.263$\pm$0.079 & 0.324$\pm$0.017   & \textbf{0.326$\pm$0.006}  & 0.292$\pm$0.006\\
\hline
\multirow{3}{*}{A3} & 5$\delta_3$ & \textbf{0.581$\pm$0.022} & 0.563$\pm$0.019  & 0.543$\pm$0.054  & 0.483$\pm$0.034 \\
& 10$\delta_3$ & 0.531$\pm$0.028  & \textbf{0.560$\pm$0.021}  &  0.526$\pm$0.037 & 0.453$\pm$0.034 \\
& 20$\delta_3$ & 0.387$\pm$0.033 & \textbf{0.414$\pm$0.011}  & 0.402$\pm$0.023 & 0.381$\pm$0.021 \\
\hline
\end{tabular}
\end{center}
\caption{CRSL vs PORS. Each cell represents the recall on test set (mean$\pm$ std) under a corresponding precision threshold. For propriety datasets, we use the precision threshold of 5$\times$, 10$\times$, 20$\times$ of the positive rate of the dataset. The best method is marked in bold for each case.}
\label{tbl:crsl_comp}
\end{table}
}

Table~\ref{tbl:crsl_comp} summarizes results on both public and propriety datasets for CRSL and PORS algorithms.
We observe that on both public and propriety datasets, PORS with \texttt{hvc-ss} outperforms CRSL on most cases. In addition, we observe that there is a positive correlation between HV (as an intermediate evaluation metric) and recall (as a final objective).

Note that CRSL needs to run on each precision threshold to produce a rule subset, while PORS algorithm only needs to run \emph{once} to generate a set of rule subsets that can be readily picked up for different precision thresholds. Such flexibility is usually appreciated in real development of rule subsets as rule mining in practice is typically an explorative and iterative process.

\textbf{Online Evaluation and Deployment}. We perform an online A/B testing for PORS algorithm (with \texttt{hvc-ss}) and CRSL~\cite{li2022adaptive} from 12/06/2023 to 12/27/2023. These two algorithms are trained on a same anti-money laundering dataset with 5,532,623 points and 717 features. The precision threshold has been set to 0.8 and the goal is to maximize recall. These two algorithms both produce rule subsets of 22 rules. 

In accordance with Alipay's non-disclosure policy, we only report the relative increase of recall rather than the actual number in this paper. PORS achieved 1.56\% increase of recall while CRSL achieved 1.49\% increase of recall; significant with $p$-value = 0 from two-tailed paired $t$-test on 59 million samples. We have deployed our PORS algorithm in our internal Fanglue system~\cite{qian2023fanglue} and it has been serving our internal users since May 2023. We have in total about 23,000 jobs up to March 2024, with an average of about 2,000 jobs per month.

\subsubsection{Rule Set Learning Using $F_\beta$ Score}
\label{sec:case_study_fbeta}
As our second case study, we focus on applications that prioritize precision over recall and we use $F_\beta$ score to select the final rule subsets for Stage 2 (which is also widely used inside Alipay). Note that for Fintech applications, a small $\beta$ is usually favored so that the metric is more biased towards precision. We compare PORS with standard greedy forward stepwise algorithm for subset selection (Greedy), both using SpectralRules to produce a initial rule set of 500 rules in Stage 1. The greedy algorithm iteratively adds one rule from the initial rule set so as to achieve the highest metric value. To avoid running into local optima, we equip Greedy with beam search of width 10.

Table~\ref{tbl:fscore_comp} summarizes results on both public and propriety datasets for Greedy and PORS algorithms. We can see that PORS with \texttt{hvc-ss} outperforms Greedy by finding rule subsets with higher $F_\beta$ on most cases. We again observe that the positive correlation between HV (as an intermediate evaluation metric) and $F_\beta$ (as a final objective). Note that Greedy needs to run multiple times (one for each $\beta$) while PORS only needs to run once and it generates a set of Pareto-optimal rule subsets. It gives more flexibility for human to quickly experiment different $\beta$s without additional efforts.

\begin{figure}[t]
\begin{center}
\includegraphics[width=0.4\textwidth]{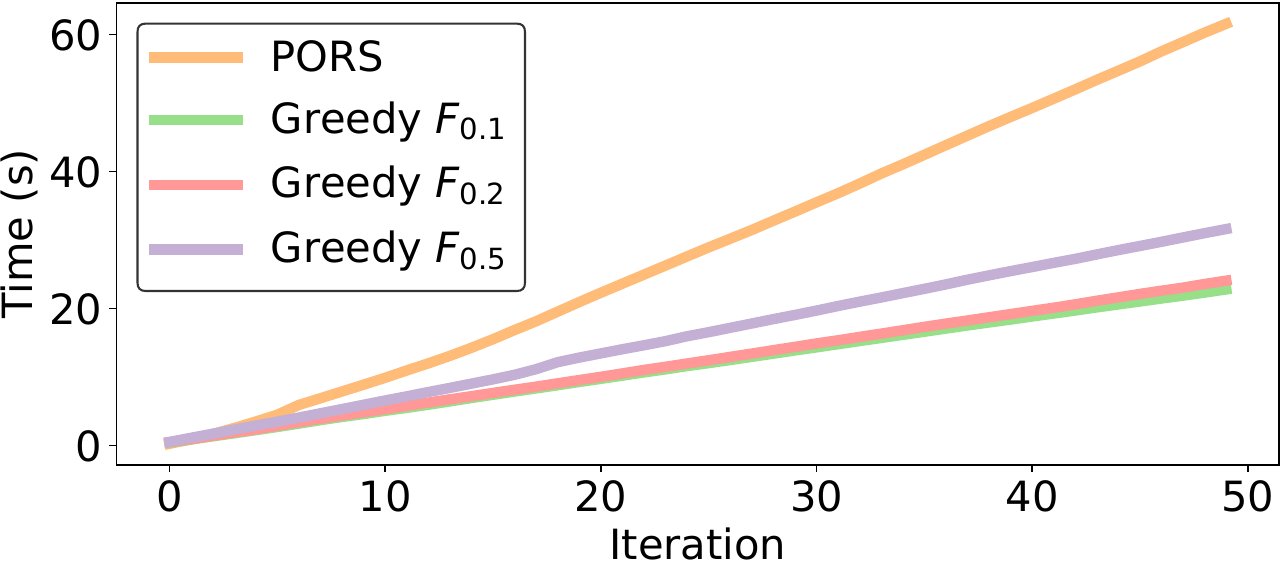}
\end{center}
\caption{Running time for Greedy and PORS on Bank dataset.}
\label{fig:fscore}
\end{figure}

We note that both Greedy and PORS adds one rule for each iteration, and therefore we can compare the running time in a relatively fair fashion. Hence, we run both Greedy and PORS (with \texttt{hvc-ss}) algorithm on a standard server with 16GB memory using 4 threads. Figure~\ref{fig:fscore} illustrates the running time for PORS and Greedy $F_{0.1}$, $F_{0.2}$, and $F_{0.5}$ on Bank dataset. As expected, $\beta$ has little impact on the running time for Greedy. Since PORS aims to generate a set of rule subsets while Greedy only focuses on one single metric, PORS needs to do more work for each iteration and we empirically observe PORS takes 2x to 3x more time than single run of Greedy for each iteration. This is acceptable in practice given the flexibility offered by PORS.

{\footnotesize
\begin{table}[t]
\begin{center}
\begin{tabular}{cccccc}
\hline
\multirow{3}{*}{Dataset} & \multirow{3}{*}{$\beta$} & \multirow{3}{*}{Greedy} & \multicolumn{3}{c}{PORS} \\
& & & \texttt{hvc-ss} & \texttt{equi-spaced} & \texttt{k-med.-j.} \\
& & & (High HV) & (Med. HV) &  (Low HV) \\
\hline\hline
\multirow{3}{*}{Default} & 0.1 & 0.743$\pm$0.039 & \textbf{0.762$\pm$0.033} & 0.747$\pm$0.022 & 0.732$\pm$0.007 \\
& 0.2 & 0.685$\pm$0.015 & 0.687$\pm$0.015 & \textbf{0.692$\pm$0.019} & 0.685$\pm$0.012 \\
& 0.5 & 0.573$\pm$0.014 & 0.578$\pm$0.008 & 0.577$\pm$0.011 & \textbf{0.582$\pm$0.014} \\
\hline
\multirow{3}{*}{Credit} & 0.1 & 0.608$\pm$0.043 & \textbf{0.616$\pm$0.035} & 0.608$\pm$0.039 & 0.611$\pm$0.034 \\
& 0.2 & 0.523$\pm$0.020 & 0.536$\pm$0.015 & \textbf{0.540$\pm$0.025} & 0.534$\pm$0.025 \\
& 0.5 & 0.431$\pm$0.011 & 0.433$\pm$0.009 & \textbf{0.434$\pm$0.007} & 0.426$\pm$0.011 \\
\hline
\multirow{3}{*}{Fraud} & 0.1 & \textbf{0.973$\pm$0.006} & 0.968$\pm$0.013 & 0.959$\pm$0.023 & 0.957$\pm$0.011 \\
& 0.2 & 0.935$\pm$0.029 & \textbf{0.950$\pm$0.007} & 0.941$\pm$0.016 & 0.946$\pm$0.008 \\
& 0.5 & 0.838$\pm$0.032 & \textbf{0.888$\pm$0.021} & 0.887$\pm$0.021 & 0.873$\pm$0.019 \\
\hline
\multirow{3}{*}{Bank} & 0.1 & 0.755$\pm$0.017 & \textbf{0.763$\pm$0.020} & 0.709$\pm$0.016 & 0.729$\pm$0.016 \\
& 0.2 & 0.652$\pm$0.017 & \textbf{0.682$\pm$0.015} & 0.666$\pm$0.018 & 0.656$\pm$0.017 \\
& 0.5 & 0.596$\pm$0.009 & \textbf{0.603$\pm$0.010} & 0.596$\pm$0.012 & 0.586$\pm$0.010 \\
\hline\hline
\multirow{3}{*}{A1} & 0.1 & 0.689$\pm$0.011 & \textbf{0.695$\pm$0.009} & 0.682$\pm$0.009 & 0.674$\pm$0.010 \\
& 0.2 & 0.592$\pm$0.010 & \textbf{0.599$\pm$0.006} & 0.599$\pm$0.009 & 0.590$\pm$0.010 \\
& 0.5 & 0.446$\pm$0.004 & \textbf{0.451$\pm$0.006} & 0.443$\pm$0.003 & 0.439$\pm$0.007 \\
\hline
\multirow{3}{*}{A2} & 0.1 & 0.543$\pm$0.045 & \textbf{0.547$\pm$0.039} & 0.543$\pm$0.044 & 0.535$\pm$0.020 \\
& 0.2 & 0.443$\pm$0.018 & \textbf{0.446$\pm$0.025} & 0.438$\pm$0.024 & 0.429$\pm$0.017 \\
& 0.5 & 0.290$\pm$0.012 & 0.290$\pm$0.012 & \textbf{0.304$\pm$0.011} & 0.289$\pm$0.009 \\
\hline
\multirow{3}{*}{A3} & 0.1 & \textbf{0.257$\pm$0.183} & 0.250$\pm$0.114 & 0.240$\pm$0.119 & 0.222$\pm$0.139 \\
& 0.2 & 0.119$\pm$0.076 & \textbf{0.127$\pm$0.062} & 0.126$\pm$0.060 & 0.119$\pm$0.068 \\
& 0.5 & 0.067$\pm$0.029 & \textbf{0.074$\pm$0.019} & 0.065$\pm$0.017 & 0.063$\pm$0.009 \\
\hline
\end{tabular}
\end{center}
\caption{Greedy vs PORS. Each cell represents the $F_\beta$ score on test set (mean$\pm$ std) under a particular $\beta$. The best method is marked in bold for each case.}
\label{tbl:fscore_comp}
\end{table}
}

\balance
\section{Conclusion}
\label{sec:conclusion}

The two-stage framework of fraud prevention decision rule set mining is usually employed in large Fintech institutions. This paper focuses on improving the flexibility and efficacy of this two-stage framework and is concerned with finding high-quality rule subsets in a bi-objective space. We first introduce a novel algorithm called SpectralRules for Stage 1 that encourages the diversity of rules. We empirically find such  diversity improves the final quality of rule subset. We also introduce an intermediate stage between Stage 1 and 2 that adopts the concept of Pareto optimality and aims to find a set of non-dominated rule subsets, which constitutes a Pareto front. This intermediate stage greatly simplifies the selection criteria and increases the flexibility for Stage 2. We propose a heuristic-based framework called PORS and we identify that the core of PORS is the problem of solution selection on the front (SSF). We provide a thorough empirical evaluation of various SSF methods on both public and proprietary datasets. On two real application scenarios within Alipay, we demonstrate the advantages of our proposed methodology compared to existing work.

\bibliographystyle{ACM-Reference-Format}
\bibliography{paper}

\appendix

{\small
\begin{table*}[!bp]
\begin{center}
\begin{tabular}{c|c||c|c|c|c|c|c|c|c|c}
    \hline
    Dataset & k & hvc-ss & hv-ss  & igd-ss & igd+-ss & k-med.-pr & k-med.-j. & equi-j. & equi-spaced & equi-dist \\
    \hline
    \hline
    \multirow{4}{*}{Bank} & 5 & 0.575$\pm$0.009 & \textbf{0.583$\pm$0.011} & 0.531$\pm$0.014 & 0.502$\pm$0.025 & 0.549$\pm$0.012 & 0.493$\pm$0.022 & 0.552$\pm$0.012 & 0.555$\pm$0.009 & 0.561$\pm$0.01 \\
    \cline{2-11}
    & 10 & \textbf{0.591$\pm$0.012} & 0.580$\pm$0.009 & 0.546$\pm$0.017 & 0.539$\pm$0.016 & 0.562$\pm$0.010 & 0.512$\pm$0.013 & 0.564$\pm$0.015 & 0.571$\pm$0.009 & 0.571$\pm$0.010 \\
    \cline{2-11}
    & 15 & \textbf{0.596$\pm$0.011} & 0.582$\pm$0.010 & 0.564$\pm$0.013 & 0.553$\pm$0.014 & 0.566$\pm$0.008 & 0.533$\pm$0.018 & 0.567$\pm$0.012 & 0.574$\pm$0.006 & 0.577$\pm$0.008 \\
    \cline{2-11}
    & 20 & \textbf{0.594$\pm$0.011} & 0.583$\pm$0.010 & 0.561$\pm$0.011 & 0.554$\pm$0.013 & 0.571$\pm$0.011 & 0.531$\pm$0.011 & 0.568$\pm$0.014 & 0.574$\pm$0.006 & 0.578$\pm$0.008 \\
    \hline
    \multirow{4}{*}{Credit} & 5 & 0.347$\pm$0.012 & \textbf{0.350$\pm$0.011} & 0.333$\pm$0.010 & 0.325$\pm$0.012 & 0.337$\pm$0.010 & 0.316$\pm$0.008 & 0.342$\pm$0.016 & 0.340$\pm$0.016 & 0.341$\pm$0.012 \\
    \cline{2-11}
    & 10 & \textbf{0.357$\pm$0.012} & 0.356$\pm$0.015 & 0.345$\pm$0.014 & 0.336$\pm$0.011 & 0.345$\pm$0.011 & 0.331$\pm$0.011 & 0.355$\pm$0.013 & 0.350$\pm$0.013 & 0.351$\pm$0.015 \\
    \cline{2-11}
    & 15 & \textbf{0.359$\pm$0.015} & 0.358$\pm$0.014 & 0.348$\pm$0.014 & 0.340$\pm$0.012 & 0.349$\pm$0.013 & 0.337$\pm$0.011 & 0.356$\pm$0.013 & 0.353$\pm$0.013 & 0.356$\pm$0.014 \\
    \cline{2-11}
    & 20 & \textbf{0.359$\pm$0.015} & 0.358$\pm$0.014 & 0.352$\pm$0.014 & 0.342$\pm$0.011 & 0.350$\pm$0.014 & 0.341$\pm$0.012 & 0.356$\pm$0.013 & 0.355$\pm$0.013 & 0.357$\pm$0.013 \\
    \hline
    \multirow{4}{*}{Default} & 5 & \textbf{0.541$\pm$0.014} & 0.532$\pm$0.013 & 0.501$\pm$0.021 & 0.509$\pm$0.022 & 0.512$\pm$0.014 & 0.495$\pm$0.019 & 0.532$\pm$0.010 & 0.529$\pm$0.012 & 0.534$\pm$0.012 \\
    \cline{2-11} 
    & 10 & \textbf{0.540$\pm$0.012} & 0.535$\pm$0.010 & 0.517$\pm$0.015 & 0.515$\pm$0.016 & 0.515$\pm$0.013 & 0.504$\pm$0.021 & 0.535$\pm$0.009 & 0.534$\pm$0.012 & 0.535$\pm$0.011 \\
    \cline{2-11} 
    & 15 & \textbf{0.536$\pm$0.012} & 0.535$\pm$0.011 & 0.524$\pm$0.016 & 0.525$\pm$0.018 & 0.523$\pm$0.015 & 0.510$\pm$0.015 & 0.535$\pm$0.010 & 0.536$\pm$0.011 & 0.536$\pm$0.010 \\
    \cline{2-11}
    & 20 & 0.533$\pm$0.011 & \textbf{0.537$\pm$0.010} & 0.525$\pm$0.012 & 0.525$\pm$0.013 & 0.527$\pm$0.014 & 0.514$\pm$0.015 & 0.534$\pm$0.012 & 0.536$\pm$0.011 & \textbf{0.537$\pm$0.010} \\
    \hline
    \multirow{4}{*}{Fraud} & 5 & \textbf{0.805$\pm$0.031} & 0.794$\pm$0.034 & 0.801$\pm$0.026 & 0.798$\pm$0.024 & 0.802$\pm$0.030 & 0.786$\pm$0.035 & 0.794$\pm$0.027 & 0.793$\pm$0.039 & 0.790$\pm$0.034 \\
    \cline{2-11}
    & 10 & \textbf{0.808$\pm$0.030} & 0.801$\pm$0.033 & 0.798$\pm$0.045 & 0.794$\pm$0.026 & 0.793$\pm$0.032 & 0.800$\pm$0.029 & 0.803$\pm$0.028 & 0.796$\pm$0.037 & 0.805$\pm$0.025 \\
    \cline{2-11}
    & 15 & 0.806$\pm$0.032 & 0.801$\pm$0.032 & \textbf{0.809$\pm$0.030} & 0.806$\pm$0.032 & 0.802$\pm$0.035 & 0.804$\pm$0.029 & 0.808$\pm$0.025 & 0.797$\pm$0.036 & 0.807$\pm$0.028 \\
    \cline{2-11}
    & 20 & 0.801$\pm$0.037 & 0.797$\pm$0.045 & 0.797$\pm$0.048 & 0.807$\pm$0.026 & \textbf{0.808$\pm$0.033} & 0.797$\pm$0.035 & 0.793$\pm$0.047 & 0.806$\pm$0.033 & 0.804$\pm$0.040 \\
    \hline
\end{tabular}
\end{center}
\caption{The hypervolume (HV) performance (mean$\pm$std) of the Pareto front for different methods on test set of each problem. Higher HV is better. SpectralRules is employed in Stage 1 to produce the initial 500 rules. The maximum round running of PORS is set to 30 to avoid early stopping, i.e., each experiment trial fully completed 30 iterations.}
\label{tbl:ssf_hv}
\end{table*}
}

{\small
\begin{table*}[!bp]
\begin{center}
\begin{tabular}{c|c||c|c|c|c|c|c|c|c|c}
    \hline
    Dataset & k & hvc-ss & hv-ss  & igd-ss & igd+-ss & k-med.-pr & k-med.-j. & equi-j. & equi-spaced & equi-dist \\
    \hline
    \hline
    \multirow{4}{*}{Bank} & 5 & 47.36$\pm$2.00 & 47.36$\pm$2.00 & 46.71$\pm$0.80 & 47.09$\pm$2.13 & 44.85$\pm$0.94 & 59.33$\pm$3.25 & 48.10$\pm$4.76 & 44.02$\pm$1.68 & 44.65$\pm$1.73 \\
    \cline{2-11}
    & 10 & 64.27$\pm$4.08 & 53.61$\pm$3.08 & 57.44$\pm$2.16 & 58.02$\pm$2.95 & 53.62$\pm$3.05 & 73.65$\pm$2.70 & 65.63$\pm$2.55 & 52.64$\pm$2.19 & 52.21$\pm$2.59 \\
    \cline{2-11}
    & 15 & 83.45$\pm$4.23 & 57.56$\pm$1.82 & 69.57$\pm$4.38 & 66.63$\pm$3.65 & 60.31$\pm$3.89 & 93.31$\pm$3.93 & 73.08$\pm$1.16 & 55.72$\pm$1.29 & 56.26$\pm$2.15 \\
    \cline{2-11}
    & 20 & 113.47$\pm$4.22 & 62.71$\pm$1.11 & 80.73$\pm$4.80 & 75.91$\pm$1.23 & 62.19$\pm$2.07 & 110.80$\pm$4.50 & 81.95$\pm$4.08 & 62.28$\pm$2.74 & 61.53$\pm$2.47 \\
    \hline
    \multirow{4}{*}{Credit} & 5 & 57.45$\pm$6.17 & 48.92$\pm$5.83 & 58.95$\pm$6.63 & 49.92$\pm$2.17 & 48.03$\pm$3.61 & 125.73$\pm$15.02 & 68.48$\pm$5.81 & 45.31$\pm$1.19 & 48.92$\pm$3.68 \\
    \cline{2-11}
    & 10 & 93.31$\pm$9.57 & 63.53$\pm$4.56 & 90.76$\pm$6.50 & 73.82$\pm$4.17 & 61.64$\pm$4.28 & 238.02$\pm$25.20 & 121.91$\pm$12.10 & 57.66$\pm$3.41 & 61.02$\pm$8.11 \\
    \cline{2-11}
    & 15 & 140.18$\pm$10.64 & 70.92$\pm$4.03 & 121.71$\pm$12.48 & 93.68$\pm$2.77 & 68.72$\pm$4.27 & 350.31$\pm$22.91 & 161.11$\pm$8.78 & 66.79$\pm$2.86 & 71.79$\pm$11.58 \\
    \cline{2-11}
    & 20 & 185.67$\pm$8.33 & 81.91$\pm$8.87 & 164.50$\pm$13.35 & 119.50$\pm$8.09 & 74.65$\pm$4.09 & 512.19$\pm$48.05 & 196.48$\pm$18.93 & 76.58$\pm$3.32 & 79.19$\pm$7.05 \\
    \hline
    \multirow{4}{*}{Default} & 5 & 49.19$\pm$2.19 & 43.57$\pm$2.35 & 49.69$\pm$2.29 & 47.44$\pm$1.78 & 41.26$\pm$1.07 & 53.80$\pm$3.85 & 57.73$\pm$2.32 & 41.82$\pm$1.03 & 43.08$\pm$1.74 \\
    \cline{2-11} 
    & 10 & 71.33$\pm$4.34 & 54.12$\pm$5.40 & 66.04$\pm$4.95 & 63.25$\pm$3.70 & 49.20$\pm$3.36 & 71.20$\pm$2.21 & 79.19$\pm$3.06 & 51.53$\pm$2.41 & 49.99$\pm$1.75 \\
    \cline{2-11} 
    & 15 & 99.13$\pm$3.36 & 60.22$\pm$4.73 & 80.24$\pm$1.89 & 79.72$\pm$6.15 & 54.43$\pm$2.38 & 93.06$\pm$6.31 & 96.18$\pm$4.66 & 56.71$\pm$2.01 & 55.31$\pm$1.72 \\
    \cline{2-11}
    & 20 & 133.09$\pm$10.86 & 67.49$\pm$5.00 & 101.47$\pm$2.29 & 98.13$\pm$3.53 & 61.73$\pm$3.17 & 116.56$\pm$4.28 & 113.25$\pm$5.85 & 61.16$\pm$1.81 & 61.50$\pm$1.45 \\
    \hline
    \multirow{4}{*}{Fraud} & 5 & 40.42$\pm$0.16 & 40.81$\pm$1.03 & 41.35$\pm$1.05 & 42.23$\pm$0.81 & 41.15$\pm$0.55 & 61.24$\pm$1.63 & 42.40$\pm$0.52 & 42.23$\pm$0.54 & 42.20$\pm$0.55 \\
    \cline{2-11}
    & 10 & 45.75$\pm$1.00 & 45.62$\pm$1.14 & 46.94$\pm$0.74 & 47.88$\pm$1.52 & 46.68$\pm$0.81 & 74.92$\pm$1.87 & 48.86$\pm$1.23 & 47.35$\pm$1.36 & 47.44$\pm$1.85 \\
    \cline{2-11}
    & 15 & 50.88$\pm$0.84 & 50.06$\pm$0.43 & 51.90$\pm$1.41 & 52.43$\pm$0.80 & 50.34$\pm$0.52 & 87.41$\pm$2.18 & 51.62$\pm$1.31 & 50.27$\pm$0.31 & 50.49$\pm$0.91 \\
    \cline{2-11}
    & 20 & 56.71$\pm$2.46 & 53.91$\pm$0.63 & 55.59$\pm$1.22 & 56.99$\pm$0.99 & 54.65$\pm$0.68 & 100.30$\pm$1.85 & 55.06$\pm$1.58 & 54.12$\pm$0.58 & 53.56$\pm$0.41 \\
    \hline
\end{tabular}
\end{center}
\caption{The running time in seconds (mean$\pm$std) for different methods on test set of each problem. SpectralRules is employed in Stage 1 to produce the initial 500 rules. The maximum round running of PORS is set to 30 to avoid early stopping, i.e., each experiment trial fully completed 30 iterations.}
\label{tbl:ssf_time}
\end{table*}
}

\section{Impact of Different Values of $k$}
\label{sec:impact_k}

To further evaluate the impact of different values of $k$ on PORS with different SSF methods, including running time and HV, we set $k$ to 5, 10, 15, 20 respectively and apply those 9 PORS algorithms on the 4 public datasets. According to previous sections, SpectralRules is employed in Stage 1 and the maximum round running of PORS is set to 30 to avoid early stopping, i.e., each experiment trial fully completed 30 iterations. All experiments are repeated 5 times.

Table~\ref{tbl:ssf_hv} shows the HV results on test set of different PORS algorithms for each $k$. Once again, we observe that PORS with \texttt{hvc-ss} as its SSF method achieves the highest HV on most cases. For all SSF methods, a larger value of $k$ usually leads to a higher HV as expected. However, once the $k$ exceeds 10, we observe the standard diminishing returns phenomenon between $k$ and HV. 

The running time of PORS is positively correlated to the value of $k$ as shown in Table~\ref{tbl:ssf_time}. Considering the trade-off between performance and efficiency, we recommend using $k = 10$ in practice.

\end{document}